\documentclass[conference]{IEEEtran}

\usepackage{cite}
\usepackage{amsmath,amssymb,amsfonts}
\usepackage{graphicx}
\usepackage{textcomp}
\usepackage{xcolor}
\usepackage{algorithm}  
\usepackage{algpseudocode}  

\makeatletter  
\def\BState{\State\hskip-\ALG@thistlm}  
\makeatother  

\floatname{algorithm}{Procedure}

\def\BibTeX{{\rm B\kern-.05em{\sc i\kern-.025em b}\kern-.08em
    T\kern-.1667em\lower.7ex\hbox{E}\kern-.125emX}}
\begin{document}
\IEEEoverridecommandlockouts

\title{Vertical federated learning based on DFP and BFGS}

\author{\IEEEauthorblockN{1\textsuperscript{st} Wenjie Song}
\IEEEauthorblockA{\textit{China University of Petroleum} \\
Qingdao, China \\
1709010220@s.upc.edu.cn}
\and
\IEEEauthorblockN{2\textsuperscript{nd} Xuan Shen}
\IEEEauthorblockA{\textit{China University of Petroleum} \\
Qingdao, China\\
1709010108@s.upc.edu.cn}
\and
\IEEEauthorblockN{3\textsuperscript{rd} Given Name Surname}
\IEEEauthorblockA{\textit{dept. name of organization (of Aff.)} \\
City, Country \\
email address}
}

\maketitle
\begin{abstract}
As data privacy is gradually valued by people, federated learning(FL) has emerged because of its potential to protect data. FL uses homomorphic encryption and differential privacy encryption on the promise of ensuring data security to realize distributed machine learning by exchanging encrypted information between different data providers.
However, there are still many problems in FL, such as the communication efficiency between the client and the server and the data is non-iid.
In order to solve the two problems mentioned above, we propose a novel vertical federated learning framework based on the DFP and the BFGS(denoted as BDFL), then apply it to logistic regression.
Finally, we perform experiments using real datasets to test efficiency of BDFL framework.
\end{abstract}
\hspace*{\fill} 

\begin{IEEEkeywords}
Federated learning, Machine learning, Non-iid data, Data privacy.
\end{IEEEkeywords}

\section{INTRODUCTION}
On the one hand, due to the emergence of the General Data Protection Regulation, more and more people are paying attention to privacy protection in machine learning. On the other hand, in real situations, more and more data island appears, making traditional machine learning difficult to achieve. Generally speaking, AI service needs data provided by users to train on a server. However, in this process, the data may come from various institutions, and although the institution wants to get a perfect model, it does not like leaking its own data.
Therefore, in order to break data island and achieve privacy protection, Google \cite{google} proposed federated learning in 2016. In FL, AI services can perform machine learning without collecting data from various institutions. FL allows the model to be trained locally and send encrypted information to the center server. Then the center server aggregates received data and send back to every client. Finally client could update parameter by themselves.
For the method of updating parameters, there are \textbf{GD}, \textbf{SGD}, \textbf{Mini-Batch SGD} methods, but these methods are all first-order accuracy. Therefore, we consider a higher-order accuracy method, the \textbf{newton} method, but in the \textbf{newton} method, the Hessian matrix may be irreversible and even if it does be a inverse matrix, it is also extremely difficult to compute it. Therefore, we consider adopting the \textbf{quasi-newton method}. Among them, \textbf{DFP} and \textbf{BFGS} are two representative algorithms. Yang \cite{quasinewton} et al. implemented \textbf{BFGS} under the algorithm architecture of logistic regression and applied it to vertical federated learning.
But in terms of communication, there are still problems. Therefore, we combined \textbf{DFP} and \textbf{BFGS} to propose a new algorithm, which is used in the logistic regression algorithm of vertical federated learning. In the end, compared to other algorithm, our algorithm can achieve better results with less communication times.
\section{RELATED WORK}
In recent years, a large number of studies on federated learning have emerged \cite{fl_1}, \cite{fl_2}, \cite{fl_3}. In its architecture, the use of gradient descent methods is common. However, the convergence of the first-order gradient descent method is lower than that of the second-order \textbf{newton} method. The calculation is very large when calculating the inverse of the Hesian matrix, so the \textbf{quasi-newton} method came into being, \textbf{BFGS} and \textbf{DFP}, as the two representative methods.
A series of works on horizontal federated learning has been proposed \cite{horizen_1},\cite{horizen_2}, each client has a part of the sample, but has all the data attributes. In vertical federated learning, each client holds part of the data attributes, and the samples are overlapped. \cite{vertical_1} suggests that logistic regression is applied under the framework of vertical federation. Yang \cite{quasinewton} and others use \textbf{L-BFGS} to implement logistic regression algorithm of vertical federated learning. It reduces communication cost. \cite{blockfl} combines federated learning with blockchain proposing BlockFL. Because of the consensus mechanism in the blockchain, BlockFL can resist attacks from malicious clients. \textbf{FedAvg} \cite{fl_2} is an iterative method that has become a universal optimization method in FL. In addition, in terms of theoretical proof, \cite{proof_1}, \cite{proof_2} gives a proof of convergence for the \textbf{FedAvg} algorithm for non-IID data. In particular, \cite{boost} offers a boosting method based on tree model SecureBoost.
Recently, \cite{li2020federated} proposes the \textbf{FedProx} algorithm on the basis of \textbf{FedAvg} by adding proximal term. \textbf{FedProx} is absolutely superior to \textbf{FedAvg} in statistical heterogeneity and system heterogeneity.

In summary, \textbf{FeaAvg} as the baseline in FL, shows bad performance in the case of statistical heterogeneity and system heterogeneity. As an improvement of \textbf{FedAvg}, \textbf{FedProx} has great performance in non-iid environments. The first-order gradient descent method in traditional machine learning has strong universality. But for FL, when the communication cost is much more than the calculation cost, a higher-precision algorithm should be selected. In other words, higher computation cost should be used in exchange for smaller communication cost.

\section{ANALYTICAL MODEL}
In this work, inspired by \textbf{BFGS} in logistic regression of vertical federated learning \cite{quasinewton}, we exlore a broader framework, \textbf{BDFL}, that is capable of managing heterogeneous federated environments when ensuring privacy security. Besides, our novel framework performs better than \textbf{BFGS} \cite{quasinewton} and \textbf{SGD} \cite{hardy2017private}.

\subsection{Logistic Regression}
In vertical federated learning, \cite{hardy2017private} realizes classic logistic regression method. Let $X \in R^{N\times T}$ be a data set containing $T$ data samples, and each instance has $N$ features. Corresponding data label is $y \in \{-1,+1\}^T$. Suppose there are two honest but curious participants party \textbf{A(host)} and  party \textbf{B(guest)}. \textbf{A} has only the characteristics of the data. And \textbf{B} has not only the characteristics, but also the label of the data. So $X^A \in R^{N_A \times T}$ is owned by \textbf{A} and $X^B \in R^{N_B \times T}$ is owned by \textbf{B}. Each party has different data characteristics, but the sample id is the same. Therefore, the goal of optimization is to train classification model to solve
\begin{equation} 
	\min\limits_{\boldsymbol  w\in R^N} \quad \frac{1}{T} \sum_{i}^{T}l(\boldsymbol w; \boldsymbol  x_i,y_i)
\end{equation} 
where $\boldsymbol w$ is the model parameters. So $\boldsymbol w=(\boldsymbol w^A,\boldsymbol w^B)$ where $\boldsymbol w^A \in R^{N_A}$ and $\boldsymbol w^B \in R^{N_B}$. Moreover $\boldsymbol x_i$ represents the feature of the i-th data instance and $y_i$ is the corresponding label. The loss function is negative log-likelihood
\begin{equation} \label{Second formula}
	l(\boldsymbol w ;\boldsymbol x_i,y_i)=log(1+exp(-y_i \boldsymbol  w^T \boldsymbol  x_i))
\end{equation}

In \cite{hardy2017private}, they use \textbf{SGD} to decrease gradient by exchanging encrypted middle information at each iteration. Party \textbf{A} and Party \textbf{B} hold vertically encrypted gradients $\boldsymbol g^A \in R^{n_A}$ and $\boldsymbol g^B \in R^{n_B}$ respectively, which can be decrypted by the third party \textbf{C}. Furthemore, to achieve secure multi-party computing, the additively homomorphic encryption is accepted. 
In the field of homomorphic encryption, a lot of works have been completed \cite{acar2017survey} \cite{zhu2020distributed}. Different computing requirements correspond to different encryption methods, such as \textbf{PHE}, \textbf{SHE}, \textbf{FHE}. After encryption, we can directly perform encrypted data with addition or multiplication operations, the value of decrypting the operation result is consistent with the result of the direct operation on the original data. That is $[\![m]\!] + [\![n]\!] = [\![m+n]\!]$ and $[\![m]\!] \cdot n = [\![m \cdot n]\!]$ with $[\![\cdot]\!]$ represent encryption method.
However, homomorphic encryption has no idea to solve exponential calculation yet. So equation (\ref{Second formula}) cannot directly apply homomorphic encryption. We consider using Taylor expansion to approximate the loss function. Fortunately, it's proposed in \cite{hardy2017private} as 
\begin{equation}
	l(\boldsymbol w;\boldsymbol x_i,y_i) \approx log2 - \frac{1}{2}y_i \boldsymbol  w^T \boldsymbol  x_i + \frac{1}{8}(\boldsymbol w^T \boldsymbol x_i)^2
\end{equation}

\subsection{Newton Method} 
The basic idea of \textbf{newton's} method is to use the first-order gradient and the second-order gradient(Hessian) at the iteration point to approximate the objective function with the quadratic function, and then use the minimum point of the quadratic model as the new iteration point. This process is repeated until the approximate minimum value that satisfies the required accuracy. The newton's method can highly approximate the optimal value and its speed is quite fast. Though it is very quickly, the calculation is extremely huge. For federated learning, this method is perfect when trading larger computational costs for smaller communication costs. 

For convenience, we mainly discuss the one-dimensional situation. For an objective function $f(\boldsymbol w)$, the problem of finding the extreme value of the function can be transformed into the derivative function $f'(\boldsymbol w)=0$, and the second-order Taylor expansion of the function $f(\boldsymbol w)$ is obtained
\begin{equation}
	f(\boldsymbol w)=f(\boldsymbol w_k)+f'(\boldsymbol w_k)(\boldsymbol w- \boldsymbol w_k)+\frac{1}{2}f''(\boldsymbol w_k)(\boldsymbol  w- \boldsymbol w_k)^2
\end{equation}
and take the derivative of the above formula and set it to 0, then
\begin{equation}
	f'(\boldsymbol w_k)+f''(\boldsymbol w_k)(\boldsymbol w- \boldsymbol w_k)=0
\end{equation}
\begin{equation}
	\boldsymbol w = \boldsymbol w_k-\frac{f'(\boldsymbol w_k)}{f''(\boldsymbol w_k)}
\end{equation}
it is further organized into the following iterative expression:
\begin{equation}
	\boldsymbol w_{k+1}=\boldsymbol w_k-\lambda H^{-1}f'(\boldsymbol w_k) \label{newton equation}
\end{equation}
where $\lambda$ represent step-size and $H$ represent Hessian.

This formula is an iterative formula of \textbf{newton} method.
But this method also has a fatal flaw, that is, in equation \eqref{newton equation}, the inverse of the Hessian matrix needs to be required. As we all know, not all matrices have inverses. And the computational complexity of the inversion operation is also very large. Therefore, there is quasi-newton methods. \textbf{BFGS} and \textbf{DFP}, approximate newton method.

\subsection{Quasi-Newton Method}\label{AA}
The central idea of the \textbf{quasi-newton} method is getting a matrix similar to the Hessian inverse without computing the inverse of Hessian. Therefore, the expression of the \textbf{quasi-newton} method is similar to equation \eqref{newton equation}, as follows
\begin{equation}
	\boldsymbol w_{k+1} = \boldsymbol w_k - \lambda C_k f'(\boldsymbol w_k) \label{quasi-newton-update-w}
\end{equation} 
where $C_k$ is the matrix used to approximate $H^{-1}$. 

In contrast, the update formula is as follows in \textbf{SGD}
\begin{equation}
\boldsymbol w_{k+1} = \boldsymbol w_k - \lambda f'(\boldsymbol w_k) \label{update-w}
\end{equation}

Different \textbf{quasi-newton} methods are inconsistent with the iterative formula of $C_k$. Therefore, we explain the iterative formula of \textbf{DFP} and \textbf{BFGS} on $C_k$ below.
\subsubsection{DFP}
\begin{equation}
	C'_{i+1} = C_i + \frac{\triangle \boldsymbol  w_i \triangle \boldsymbol  w_i^T}{\triangle \boldsymbol  w_i^T \triangle \boldsymbol  g_i} - \frac{(C_i \triangle \boldsymbol  g_i)(C_i \triangle \boldsymbol  g_i)^T}{\triangle \boldsymbol  g_i^T C_i \triangle \boldsymbol g_i} \label{dfp}
\end{equation}
\subsubsection{BFGS}
\begin{equation}
	C''_{i+1} = (I-\frac{\triangle \boldsymbol  w_i \triangle \boldsymbol  g_i^T}{\triangle \boldsymbol  g_i^T \triangle \boldsymbol  w_i})C_i(I-\frac{\triangle \boldsymbol  g_i \triangle \boldsymbol  w_i^T}{\triangle \boldsymbol  g_i^T \triangle \boldsymbol  w_i}) + \frac{\triangle \boldsymbol  w_i \triangle \boldsymbol  w_i^T}{\triangle \boldsymbol  g_i^T \triangle \boldsymbol  w_i} \label{bfgs}
\end{equation}
\subsubsection{BDFL}
\begin{equation}
	C_{i+1} = \alpha C'_{i+1} + (1-\alpha) C''_{i+1} \label{quasi-newton-compute-C}
\end{equation} 

In equation \eqref{dfp} and equation \eqref{bfgs}, $\triangle \boldsymbol  w_i =  \boldsymbol w_{i+1} -  \boldsymbol w_i ,\triangle  \boldsymbol g_i =  \boldsymbol g_{i+1}- \boldsymbol g_i$. In equation \eqref{quasi-newton-compute-C}, $\alpha$ is a number with no limits.

\subsection{Compute and Exchange information}

\begin{figure}[H]
	\centering
	{\includegraphics[width=7cm,height=5cm]{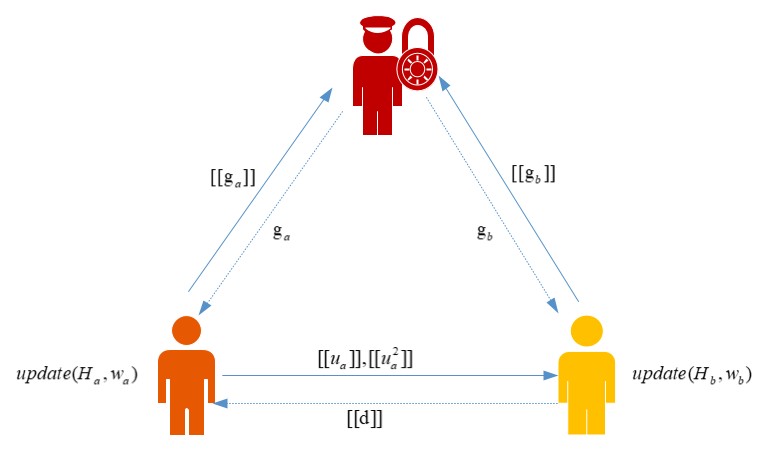}}
	\caption{Information exchange between parties}
	\label{fig-workflow}
\end{figure}

The gradient and the Hessian of Taylor loss of the i-th data sample are given by $ \boldsymbol g_i = \nabla l( \boldsymbol w; \boldsymbol x_i,y_i) \approx (\frac{1}{4} \boldsymbol w^T \boldsymbol x_i-\frac{1}{2}y_i) \boldsymbol x_i$, $H = \nabla^2l( \boldsymbol w; \boldsymbol x_i,y_i) \approx \frac{1}{4} \boldsymbol x_i \boldsymbol x_i^T$ respectively.
For convenience, we calculate the intermediate variable $ \boldsymbol w^T \boldsymbol x$ and express it
\begin{equation}
 \boldsymbol u =  \boldsymbol w^T \boldsymbol x_i	\label{compute-u}
\end{equation}
\subsubsection{Compute Gradient and Loss}
First, after initializing $w$ and $C$, both parties \textbf{A} and \textbf{B} calculate $u_a$ and $u_b$. After the calculation is completed, party \textbf{A} sends $[\![ \boldsymbol u_a]\!]$ and $[\![ \boldsymbol u_a^2]\!]$ to \textbf{B}. Next, \textbf{B} calculates $[\![loss]\!]$, $[\![ \boldsymbol d]\!]$ according to formula \eqref{compute-loss} and \eqref{compute-d} and then sends $[\![ \boldsymbol d]\!]$ to \textbf{A}. Then according to the equation \eqref{compute-g}, \textbf{A} calculates $[\![ \boldsymbol g_a]\!]$ and B calculates $[\![ \boldsymbol g_b]\!]$.
\begin{equation}
[\![ \boldsymbol d]\!] = \frac{1}{4}([\![ \boldsymbol u_A[i]]\!] + [\![ \boldsymbol u_B[i]]\!] - 2[\![y_i]\!]) \label{compute-d}
\end{equation}
\begin{equation}
[\![ \boldsymbol g]\!] \approx \frac{1}{N} \sum_{i \in N}[\![d_i]\!]x_i \label{compute-g}
\end{equation}
\begin{equation}
\begin{aligned}
\- [\![loss]\!] \approx \frac{1}{N} \sum_{i \in N} log2 - \frac{1}{2}y_i([\![ \boldsymbol u_A[i]]\!] + [\![ \boldsymbol u_B[i]]\!]) + \\  \frac{1}{8}([\![ \boldsymbol u_A^2[i]]\!]+2 \boldsymbol u_B[i][\![ \boldsymbol u_A[i]]\!] + [\![ \boldsymbol u_B^2[i]]\!] ) \label{compute-loss}
\end{aligned}
\end{equation}

\subsubsection{Send Encrypted Information And Return}
\textbf{B} sends the calculated $[\![loss]\!]$ to \textbf{C}. And \textbf{C} decrypts it and displays the results. Then \textbf{A}\&\textbf{B} send $[\![ \boldsymbol g_a]\!],[\![ \boldsymbol g_b]\!]$ to \textbf{C}. After decrypting the gradient, \textbf{C} sends back the respective gradient plaintext.
\subsubsection{Update Hessian and $ \boldsymbol w$}
After both \textbf{A} and \textbf{B} have received their respective gradients, they first update their $C_k$. Later, update $ \boldsymbol w$ using the equation \eqref{quasi-newton-update-w}.
\subsubsection{Check}
Party \textbf{A}\&\textbf{B} check whether $w$ has reached convergence. If both of them converge, then output $w$, if one of them does not converge, continue the loop.

\begin{algorithm}[H]  
	\caption{Basic Logistic Regression In Vertical FL}  
	\textbf{Input} : $w_0^A,w_0^B,X_A,X_B,Y_B,E,\lambda$ \\
	\textbf{Output} : $w^A,w^B$ \\ 
	\textbf{Party C}: Generated public key and private key \\
	\textbf{Party C}: Send private key to A and B
	\begin{algorithmic}[1]
		\For{each round $k=1,..,E$}  
		\State \textbf{Party A}: Compute $u_a,u_a^2$ as equation \eqref{compute-u}
		\State \textbf{Party A}: Send $[\![u_a]\!],[\![u_a^2]\!]$ to B.
		\State \textbf{Party B}: Compute $u_b,u_b^2$ as equation \eqref{compute-u} 
		\State \textbf{Party B}: Compute $[\![loss]\!]$ as equation  \eqref{compute-loss}
		\State \textbf{Party B}: Compute $[\![d]\!]$ as equation \eqref{compute-d} and send to A
		\State \textbf{Party A}: Compute $[\![g_A]\!]$ as equation \eqref{compute-g}
		\State \textbf{Party B}: Compute  $[\![g_B]\!]$ as equation \eqref{compute-g}
		\State \textbf{Party A\&B}: Send  $[\![g_A]\!],[\![g_B]\!]$ to C
		\State \textbf{Party C}: Decrypted $[\![g_A]\!],[\![g_B]\!]$ and send back
		\State \textbf{Party A\&B}: Update $w$ as equation \eqref{update-w}
		\EndFor  
	\end{algorithmic}  
\end{algorithm}  
\begin{algorithm}[H]  
	\caption{BDFL Framework}  
	\textbf{Input} : $w_0^A,w_0^B,X_A,X_B,Y_B,C_0^A,C_0^B,E,\lambda $ \\
	\textbf{Output} : $w^A,w^B$ \\ 
	\textbf{Party C}: Generated public key and private key \\
	\textbf{Party C}: Send private key to A and B
	\begin{algorithmic}[1]
		\For{each round $k=1,..,E$}  
		\State \textbf{Party A}: Compute $u_a,u_a^2$ as equation \eqref{compute-u}
		\State \textbf{Party A}: Send $[\![u_a]\!],[\![u_a^2]\!]$ to B.
		\State \textbf{Party B}: Compute $u_b,u_b^2$ as equation \eqref{compute-u} 
		\State \textbf{Party B}: Compute $[\![loss]\!]$ as equation  \eqref{compute-loss}
		\State \textbf{Party B}: Compute $[\![d]\!]$ as equation \eqref{compute-d} and send to A
		\State \textbf{Party A}: Compute $[\![g_A]\!]$ as equation \eqref{compute-g}
		\State \textbf{Party B}: Compute  $[\![g_B]\!]$ as equation \eqref{compute-g}
		\State \textbf{Party A\&B}: Send  $[\![g_A]\!],[\![g_B]\!]$ to C
		\State \textbf{Party C}: Decrypted $[\![g_A]\!],[\![g_B]\!]$ and send back
		\If {$k!=1$}
			\State \textbf{Party A\&B}: Update separately $C$ as equation \eqref{quasi-newton-compute-C} 
		\EndIf
		\State \textbf{Party A\&B}: Update $w$ as equation \eqref{quasi-newton-update-w}
		\EndFor  
	\end{algorithmic}  
\end{algorithm}

\section{PERFORMANCE EVALUEATION}
Our numerical experiment has two parts. In both of the experiments, we select 80\% of the data for training and check the training loss. The remaining 20\% is used as the test dataset to check the generalization ability of the model.

\subsection{Compare Quasi-Newton with SGD}
The first part is to compare \textbf{SGD} and \textbf{quasi-Newton}. It is applied to credit card dataset, which consists of 30000 instances and each instances holds 23 features. So, we shuffle the order of the instances. Party \textbf{A} holds 12 features, and Party \textbf{B} holds remaining 11 features and corresponding target. By using the two \textbf{quasi-newton} methods of \textbf{DFP} and \textbf{BFGS}, it is compared with the \textbf{SGD} method.
\begin{figure}[H]
	\centering
	{\includegraphics[width=7cm,height=5cm]{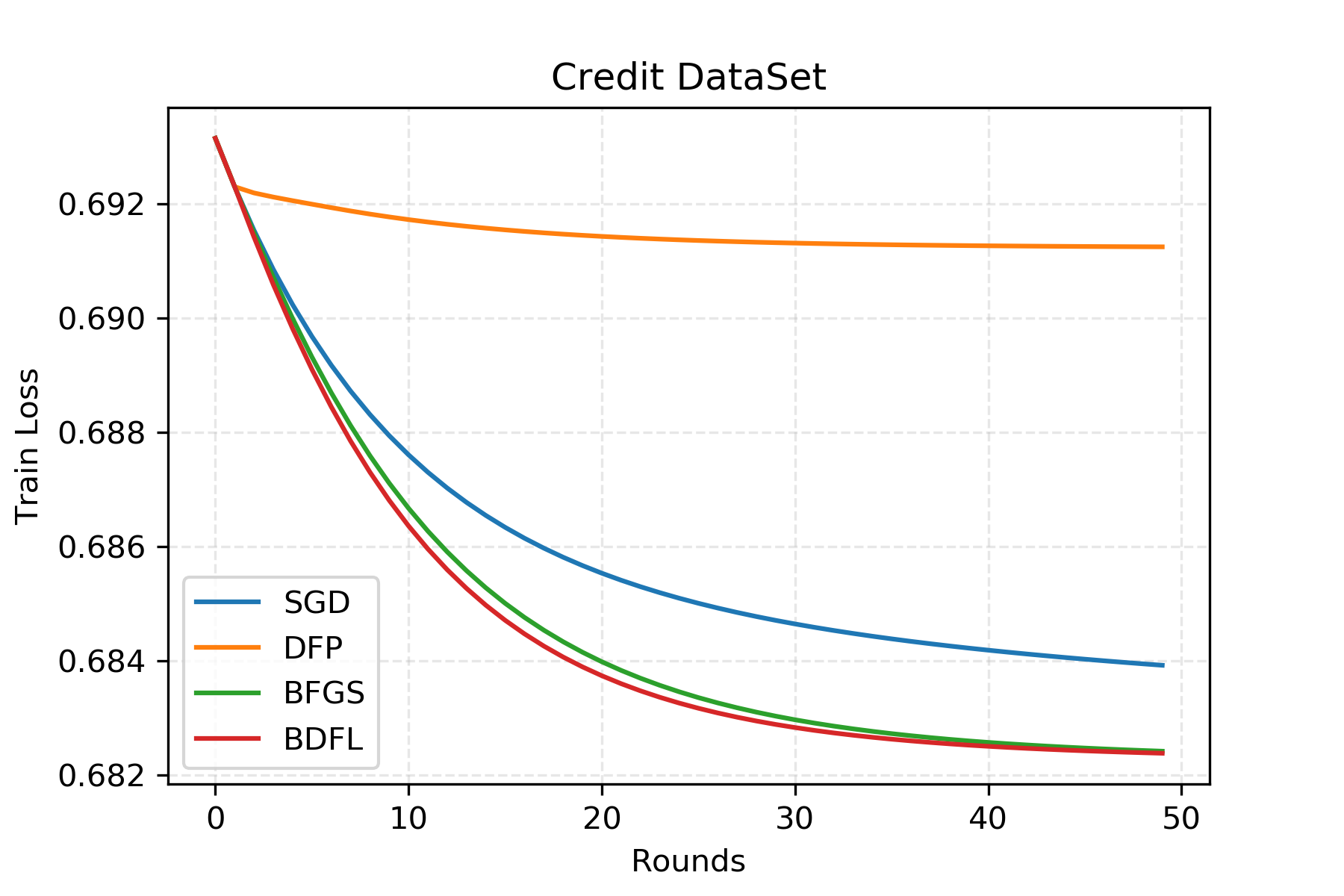}}
	\caption{Training loss in Credit Card experiment. Compare Quasi-newton method with SGD. All of them use leanring-rate decay of 0.06 per round.}
	\label{fig-credit}
\end{figure}

\subsection{Compare Quasi-Newton with BDFL}
In the second part, to go further, we use \textbf{BDFL}(we proposed) and \textbf{quasi-newton} method to compare. Using the breast cancer dataset, which has 569 instances, 30 atrributes and label. Because there are 20\% test dataset, so split them to $X_A \in R^{455 \times 20}$, $X_B \in R^{455 \times 10}$ and $Y_B \in R^{455\times1}$. The attribute index held by Party \textbf{A} are from 10-29, and those held by Party \textbf{B} are from 0-9.
\begin{figure}[H]
	\centering
	{\includegraphics[width=7cm,height=5cm]{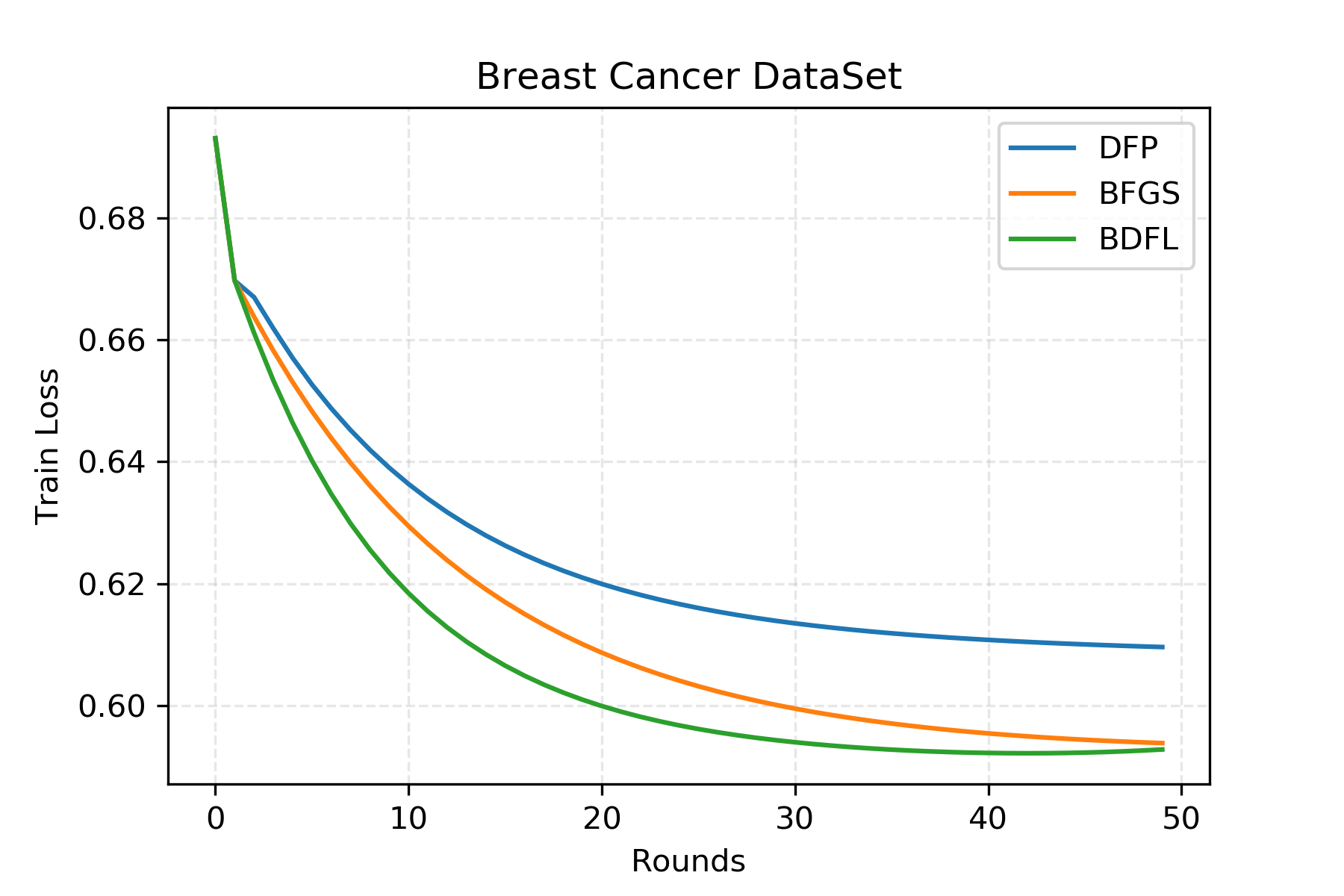}}
	\caption{Training loss in Breast Cancer experiment. Compare BFGS and DFP with BDFL. All of them use leanring-rate decay of 0.05 per round.}
	\label{fig-brest}
\end{figure}
In figure \ref{fig-credit}, it is clear that \textbf{BFGS} is much faster than \textbf{SGD}. In figure \ref{fig-brest}, it shows \textbf{BDFL} is better than both \textbf{DFP} and \textbf{BFGS}. What is more, we run every model in test dataset.
\begin{table}[H]
	\centering
	\caption{The Test Accuracy}
	\begin{tabular}{c|c|c}
		\hline
		\textbf{Method} & \textbf{\textit{Credit Card}} & \textbf{\textit{Breast Cancer}}  \\ \hline
		SGD & 90.90\% & 86.26\%  \\
		DFP & 94.41\% & 85.57\% \\
		BFGS & 95.10\% & 91.29\%  \\
		BDFL & -- & 91.35\%  \\ \hline
	\end{tabular}
\end{table}

\section{CONCLUSIONS}
In this article, we use the quasi-newton method to replace the gradient descent method on the purpose of exchanging a larger amount of calculation for a smaller communication cost. In addition, we make improvements on the original basis of the quasi-newton. A novel framework, named \textbf{BDFL}, is proposed under vertical federated learning. Logistic regression is applied to the \textbf{BDFL} framework, which is used to test actual dataset. And the experiments have shown that \textbf{BDFL} can meet the following two premises for multi-party modeling:
\begin{enumerate}
	\item Ensure data privacy would not leak;
	\item One of them has only data but no labels. The other has data and label.
\end{enumerate}
And the convergence speed and accuracy of the model are also better than traditional methods. 

But our model still has some problems, such as the convergence speed did not meet our expectations, the amount of calculation is too large, etc. We will continue to study in future work.

\bibliographystyle{paper}
\bibliography{paper}
\end{document}